\PassOptionsToPackage{unicode}{hyperref}
\PassOptionsToPackage{hyphens}{url}
\PassOptionsToPackage{dvipsnames,svgnames,x11names}{xcolor}
\documentclass[
  11pt,
  a4paper,
]{article}
\usepackage{xcolor}
\usepackage[margin=1in]{geometry}
\usepackage{amsmath,amssymb}
\usepackage{iftex}
\ifPDFTeX
  \usepackage[T1]{fontenc}
  \usepackage[utf8]{inputenc}
  \usepackage{textcomp} 
\else 
  \usepackage{unicode-math} 
  \defaultfontfeatures{Scale=MatchLowercase}
  \defaultfontfeatures[\rmfamily]{Ligatures=TeX,Scale=1}
\fi
\usepackage{lmodern}
\ifPDFTeX\else
\fi
\IfFileExists{upquote.sty}{\usepackage{upquote}}{}
\IfFileExists{microtype.sty}{
  \usepackage[]{microtype}
  \UseMicrotypeSet[protrusion]{basicmath} 
}{}
\makeatletter
\@ifundefined{KOMAClassName}{
  \IfFileExists{parskip.sty}{%
    \usepackage{parskip}
  }{
    \setlength{\parindent}{0pt}
    \setlength{\parskip}{6pt plus 2pt minus 1pt}}
}{
  \KOMAoptions{parskip=half}}
\makeatother
\usepackage{longtable,booktabs,array}
\usepackage{placeins} 
\usepackage{calc} 
\usepackage{etoolbox}
\makeatletter
\patchcmd\longtable{\par}{\if@noskipsec\mbox{}\fi\par}{}{}
\makeatother
\IfFileExists{footnotehyper.sty}{\usepackage{footnotehyper}}{\usepackage{footnote}}
\makesavenoteenv{longtable}
\usepackage{graphicx}
\makeatletter
\newsavebox\pandoc@box
\newcommand*\pandocbounded[1]{
  \sbox\pandoc@box{#1}%
  \Gscale@div\@tempa{\textheight}{\dimexpr\ht\pandoc@box+\dp\pandoc@box\relax}%
  \Gscale@div\@tempb{\linewidth}{\wd\pandoc@box}%
  \ifdim\@tempb\p@<\@tempa\p@\let\@tempa\@tempb\fi
  \ifdim\@tempa\p@<\p@\scalebox{\@tempa}{\usebox\pandoc@box}%
  \else\usebox{\pandoc@box}%
  \fi%
}
\def\fps@figure{htbp}
\makeatother
\providecommand{\tightlist}{%
  \setlength{\itemsep}{0pt}\setlength{\parskip}{0pt}}
\usepackage{bookmark}
\IfFileExists{xurl.sty}{\usepackage{xurl}}{} 
\hypersetup{
  colorlinks=true,
  linkcolor={Maroon},
  filecolor={Maroon},
  citecolor={Blue},
  urlcolor={Blue},
  pdfcreator={LaTeX via pandoc}}

\title{A Held-Out Transition-Pair Falsifier for Long-Horizon Non-Abelian State Tracking}
\author{\textbf{Jeonghoon Lee} \\ Attractor Dynamics \\ \texttt{jeonghoon@attractordynamics.ai}}
\date{}

\begin{document}

\maketitle

\begin{abstract}

State tracking exposes a sharp limitation of sequence models: the
relevant signal is often not a summary of observed tokens, but an
ordered latent state that evolves through non-commutative
transformations. We introduce a held-out transition-pair falsifier for
finite non-Abelian group tracking. The protocol forbids selected ordered
generator pairs during training and requires the same local patterns
during evaluation, blocking one direct local-transition memorization
pathway. In a controlled \(S_3 \times S_3\) benchmark, a projected
recurrent state model trained only on length-8 sequences produces
error-free final-state predictions (perfect 250/250 per horizon)
through evaluation horizons up to \texttt{1,048,576} tokens across
five seeds. Matched native-readout
baselines, including bag, GRU, and a single-configuration structured
state-space model, remain near floor under the same protocol.
Projection-matched GRU, structured SSM, and bag baselines equipped with
analogous finite-group prototype readouts also remain near chance under
the same split. Mechanism
diagnostics show that hard projection coincides with low homomorphism
error, low state-consistency drift, and non-trivial commutator
separation, while softened projection collapses final-state accuracy.
Clean-split audits verify zero verbatim reduced-word overlap and zero
structural-template overlap between training and evaluation partitions.
The evidence is scoped to this controlled finite-group falsifier rather
than to a general architecture ranking. Within that regime, explicit
projected non-commutative state composition acts as a useful inductive
bias for long-horizon hidden-state tracking.

\end{abstract}

\begin{center}\rule{0.5\linewidth}{0.5pt}\end{center}

\section{Introduction}\label{introduction}

Long-context systems are increasingly used as agents, tool users, and
workflow controllers. In such settings, failure is often not a missing
token but a corrupted hidden state: the system loses track of what has
already been done, which branch of a process it is in, or which latent
condition is currently true. A benchmark that can be solved by
local-template memorization does not test this failure mode.

Recent sequence-model evaluation increasingly emphasizes token-level
prediction quality, long-context retrieval, or in-context learning.
These framings reward models that summarize observed tokens well. They
underweight a different family of computational requirements: tasks
whose relevant signal is not a summary of any visible token sequence,
but an \textbf{ordered latent state} that evolves through composition
operations that may not commute.
 
Finite non-Abelian group tracking is a canonical controlled example.
Given a
sequence of generator symbols \((a_{t_1}, a_{t_2}, \ldots, a_{t_L})\)
drawn from a group \texttt{G}, the target is the accumulated product
\(H_L = a_{t_1} \cdot a_{t_2} \cdot \ldots \cdot a_{t_L}\). When
\texttt{G} is non-Abelian, the order matters:
\(a \cdot b \neq b \cdot a\) for at least one pair. A correct model must
therefore preserve order in its internal state across the full sequence.

Standard length-extrapolation evaluation of such tasks is vulnerable to
a subtle confound. A model that has memorized local transition patterns
observed during training can produce correct-looking outputs at longer
horizons by interpolating from those patterns, without performing
genuine non-commutative state composition. The model may appear to
extrapolate in length while in fact relying on observed
\((a_i, a_j) \to \text{next-state}\) transitions.

\textbf{A model can appear to extrapolate in length while still relying
on local transition patterns seen during training. Our goal is to remove
that path.}

This paper makes three contributions:

\begin{enumerate}
\def\labelenumi{\arabic{enumi}.}
\item
  \textbf{A held-out transition-pair falsifier.} We define a protocol
  that forbids one or more specified ordered generator-pairs from any
  training sequence, and requires those same pairs to occur in every
  evaluation sequence. Under this split, any model that solves the task
  by memorizing the specific local transition templates excluded from
  training must fail at evaluation: the required ordered pair was
  excluded from training, so no observed template covers it. Baseline
  failure under this protocol supports the interpretation that the
  direct local-template pathway has been blocked.
\item
  \textbf{A projected recurrent state model interface.} We describe a
  class of sequence models that maintain a continuous-valued
  non-commutative recurrent hidden state and produce symbolic group
  elements via a temperature-controlled projection onto the target
  finite group. We show that, under the falsifier protocol, the
  hard-projected variant of this model class preserves exact final-state
  accuracy through evaluation horizons up to approximately \(10^5\)
  times the training horizon on a worked \(S_3 \times S_3\) benchmark.
\item
  \textbf{Mechanism diagnostics under projection temperature.} We report
  a four-axis diagnostic family: final-token accuracy, exact
  homomorphism error, state-consistency drift, and commutator gap.
  Across a sweep of projection temperatures, these diagnostics identify a coherent
  boundary at which the model's representation departs from
  group-homomorphic behavior.
\end{enumerate}

A complementary clean-split overlap audit verifies that the training and
evaluation partitions of the data are non-trivially distinct under both
verbatim reduced-word and structural-template criteria.

The result is intentionally narrow. The paper contributes a falsifier
and evidence inside that falsifier, rather than a universal ranking of
sequence-model families. Concurrent work by Sung~{[}8{]} reports a
related non-Abelian length-extrapolation result on $S_{10}$ variable
binding using a different mechanism. The contribution of this paper is
accordingly narrow but concrete: under a protocol that blocks one direct
local-transition memorization pathway, explicit projected
non-commutative state composition yields exact final-state predictions at
evaluation horizons up to roughly \(10^5\) times the training length on
the controlled \(S_3 \times S_3\) benchmark.

\begin{center}\rule{0.5\linewidth}{0.5pt}\end{center}

\section{Related Work}\label{related-work}

\subsection{State tracking and sequence-model expressivity}\label{state-tracking-and-sequence-model-expressivity}

Non-commutative group composition has long served as a diagnostic of
computational expressivity. Barrington's theorem {[}1{]} established
that the word problem for finite non-Abelian groups is complete for the
circuit-complexity class \(NC^1\), and Krohn--Rhodes decomposition
{[}2{]} factored finite-state computation into group and aperiodic
primitives. Modern theoretical work on neural sequence architectures has
connected these algebraic facts to architectural limits. Merrill, Petty,
and Sabharwal {[}3{]} analyze classes of state-space sequence models and
relate their expressivity to low-depth circuit classes, motivating
finite group state tracking as a diagnostic. Shakerinava et al.~{[}4{]}
study input-dependent complex-valued diagonal SSMs and show that a
single-layer DCD SSM cannot track any non-Abelian group at finite
precision; more generally, a \texttt{k}-layer DCD SSM can express group
state tracking if and only if the group admits a length-\texttt{k}
subnormal series with Abelian factors. Ebrahimi et al.~{[}5{]}
empirically compare Transformers and recurrent models on in-distribution
state-tracking tasks, finding that Transformers require substantially
more training data as state-space size and sequence length grow and
exhibit negligible or even detrimental sharing of learned state-tracking
mechanisms across sequence lengths.

Prior work in this line turns state tracking into a diagnostic of
inductive bias. Our contribution is complementary: a stricter empirical
falsifier (the held-out transition-pair split) and a constructive
model-side result within the regime that the falsifier exposes. The
theoretical separations above motivate the benchmark family; the
present paper tests a particular empirical failure mode inside it.

Recent constructive work also seeks to recover state-tracking capability
by enriching recurrent or state-space transitions. Terzić et al.~{[}6{]}
propose PD-SSM, a structured sparse SSM whose transition matrix is
parameterized as a product of a column-one-hot matrix \texttt{P} and a
complex-valued diagonal matrix \texttt{D}, enabling finite-state
automaton tracking with efficient parallel scans. Mishra et al.~{[}7{]}
revisit nonlinear recurrent models with matrix-valued hidden states and
report state-tracking generalization together with scalable
language-modeling results. The present paper is narrower than these
architecture papers: it isolates a held-out transition-pair falsifier
and reports a projected-readout result on one controlled finite-group
benchmark.

\subsection{Non-commutative group tasks as controlled benchmarks}\label{non-commutative-group-tasks-as-controlled-benchmarks}

Permutation groups have repeatedly been used as controlled benchmarks
for sequence-model state tracking, both at the level of \(S_n\) for
small \texttt{n} and at the level of product groups exposing
direct-product structure. We use \(S_3 \times S_3\) as a compact
controlled setting under our generator-and-split design. The choice is
not motivated by a claim that \(S_3 \times S_3\) is in any sense the
smallest meaningful target. Rather, it simultaneously exposes
non-commutativity, held-out local-template generalization (in the
generator-pair embodiment used here), non-trivial commutator structure,
and same-multiset different-product distinguishability, the four
properties our protocol and diagnostics rely on, within a state space
(order 36) small enough to admit transparent statistical analysis. The
protocol extends naturally to other product non-Abelian groups,
including \(S_3 \times S_5\), \(S_5 \times S_5\), dihedral products, and
direct products containing simple non-Abelian factors. The main
controlled result of this paper is restricted to \(S_3 \times S_3\);
Appendix F additionally reports a preliminary \(S_5\) stress test, but
we do not make \(S_5\) the main claim.

\subsection{Projection, symbolic readout, and mechanism diagnostics}\label{projection-symbolic-readout-and-mechanism-diagnostics}

Existing recurrent and state-space models map their hidden state to a
continuous distribution over a class-label set, to a continuous
regression target, or to a discrete output via softmax over a
vocabulary. To the best of our knowledge, prior work has not described a
readout that \textbf{projects a continuous-valued non-commutative
recurrent hidden state onto a target finite group} to produce a symbolic
accumulated group element, with a tunable softness parameter that admits
coherent mechanism diagnostics. The projection operator we describe
(Section 4) makes the model's symbolic output explicit and admits the
diagnostic family described in Section 5.3. The four diagnostics are
final-token accuracy, exact homomorphism error, state-consistency drift,
and commutator gap. Together, they probe whether the model's
representation behaves approximately as a group homomorphism under the
hard-projection regime and how that behavior degrades as projection
softens.

\subsection{Holonomic and gauge-based non-Abelian state tracking}\label{holonomic-gauge-prior-art}

Concurrent work by Sung~[8] proposes the \emph{Holonomic Network}, which
performs non-Abelian state tracking by maintaining a hidden state on the
orthogonal manifold $SO(N)$ through an input-dependent multiplicative
update of the form $h_t = U(x_t)h_{t-1}$, where
$U(x_t) = \exp(\mathcal{A}(x_t))$ with
$\mathcal{A}(x_t) \in \mathfrak{so}(N)$. The final state is obtained as
the path-ordered product (holonomy), motivated by an effective
Chern--Simons gauge theory. On an $S_{10}$ variable-binding task, the
model reports perfect generalization when trained on sequences of length
up to 50 and evaluated at length 5000.

We view Holonomic Networks as closely related prior work within the
broader family of non-commutative recurrent architectures. Both
approaches rely on structured, order-preserving state updates on a
manifold. However, the present paper differs in several concrete
respects. First, we introduce a held-out transition-pair falsifier that
explicitly removes specific local transition templates from training
while requiring them at evaluation; Sung's experiments focus primarily
on length extrapolation rather than this form of local-template blocking.
Second, we report projection-matched baselines (GRU, structured SSM, and
bag) equipped with the same finite-group prototype readout, evaluated
under the identical held-out split. Third, our evaluation reaches
horizons up to 1,048,576 tokens, more than two hundred times longer than
the 5,000-token horizon reported in~[8]. Fourth, our mechanism uses a
continuous non-commutative state followed by an explicit projection
operator $\pi : S \to G$ onto a finite group at readout, whereas the
Holonomic Network maintains its state directly on $SO(N)$ without an
explicit finite-group projection step.

Scan parallelism arising from associativity is not a distinguishing
feature here, as Holonomic Networks also exploit this property. The
contribution of this work lies instead in the held-out transition-pair
falsifier, the use of projection-matched baselines, the explicit
finite-group readout, and error-free million-token evaluation under that
falsifier.

\begin{center}\rule{0.5\linewidth}{0.5pt}\end{center}

\section{Problem Setting}\label{problem-setting}

\subsection{Target group}\label{target-group}

We take

\[ G = S_3 \times S_3, \quad |G| = 36, \quad \Sigma = \{a_0, a_1, a_2, a_3\}. \]

where \texttt{a\_0,\ a\_1} generate the first \texttt{S\_3} factor and
\texttt{a\_2,\ a\_3} generate the second \texttt{S\_3} factor.
\texttt{G} is non-Abelian: each factor \texttt{S\_3} is non-Abelian, so
the direct product contains non-commuting pairs (e.g., pairs within the
first factor or pairs within the second factor).

\subsection{Update sequence}\label{update-sequence}

An \textbf{update sequence} of length \texttt{L} is a finite ordered
tuple

\[ w = (a_{t_1}, a_{t_2}, \ldots, a_{t_L}), \quad a_{t_i} \in \Sigma. \]

The \textbf{accumulated group state} is the cumulative product

\[ H_L(w) = a_{t_1} \cdot a_{t_2} \cdot \ldots \cdot a_{t_L} \in G. \]

evaluated under the group operation. The task is to predict
\texttt{H\_L(w)} exactly, as a symbolic group element.

\subsection{Held-out transition-pair split}\label{held-out-transition-pair-split}

Let

\[ P_{\text{forbid}} = \{(a_0, a_2), (a_2, a_0)\} \subset \Sigma \times \Sigma. \]

\textbf{Training sequences} are generated such that no training sequence
contains any pair $(a_i, a_{i+1}) \in P_forbid$ as
consecutive generators. \textbf{Evaluation sequences} are generated such
that every evaluation sequence contains at least one occurrence of each
pair in

\[ P_{\text{require}} = P_{\text{forbid}} = \{(a_0, a_2), (a_2, a_0)\}. \]

The two ordered pairs \texttt{(a\_0,\ a\_2)} and \texttt{(a\_2,\ a\_0)}
are distinct \textbf{as local transition templates} even when the
corresponding generators act on different factors of the product group.
In the present generator convention, \texttt{a\_0} belongs to the first
\texttt{S\_3} factor and \texttt{a\_2} to the second; the two factors
commute element-wise as a property of the direct product, so the two
adjacent two-token products coincide as group elements. The purpose of
holding out these specific ordered templates is therefore not to assert
that the two adjacent two-token products are themselves non-commuting,
but to \textbf{remove a specific local transition template from training
while requiring it at evaluation}. Non-commutativity of the target task
is probed at the full-sequence level (where order across many positions
determines \texttt{H\_L}) and explicitly by the commutator and
same-multiset different-product diagnostics in Gate C and Gate E.

\subsection{Why this is a falsifier}\label{why-this-is-a-falsifier}

The split functions as a falsifier in the following sense. Any model
that predicts \texttt{H\_L} by interpolating from observed local
transition templates
$(a_i, a_j) \to \text{next-state}$
must fail under this split: the required ordered template never appears
in training, so no observed template covers it. Therefore baseline
failure under this protocol is positive evidence that the protocol
blocks \textbf{one direct local-transition memorization pathway}. The
protocol does not, by itself, foreclose every conceivable memorization
or interpolation strategy; it forecloses the most direct one and exposes
whether a candidate model can succeed without it.

Conversely, success under this protocol, particularly at evaluation
horizons much longer than the training horizon, is evidence consistent
with non-commutative state composition beyond the direct
local-transition pathway. The full-sequence non-commutativity is what
distinguishes correct prediction of \texttt{H\_L} over million-token
horizons from any constant or direct template-interpolation strategy.

\subsection{Relation to a broader transition-pattern protocol}\label{relation-to-a-broader-transition-pattern-protocol}

The present paper reports the \textbf{update-pair embodiment}
(\texttt{k\ =\ 2}) of a more general held-out transition-pattern
protocol in which \texttt{P\_forbid} and \texttt{P\_require} may consist
of ordered generator-tuples of arbitrary length $k \geq 2$,
reversed-order templates, commutator templates, inverse-cancellation
templates, reduced-word templates, or same-multiset different-product
templates. We restrict the present empirical scope to the
\texttt{k\ =\ 2} update-pair case to keep the falsification reading
direct and the statistical analysis transparent. Extension to
higher-order and structural patterns is straightforward in principle.

\begin{center}\rule{0.5\linewidth}{0.5pt}\end{center}

\section{Model Interface}\label{model-interface}

We describe the model interface in a deliberately carrier-agnostic form.
The state-tracking benchmark and falsifier protocol are defined
independently of any particular continuous carrier.

\subsection{Projected recurrent state model}\label{projected-recurrent-state-model}

A \textbf{projected recurrent state model} consists of (i) a
continuous-valued recurrent hidden state, (ii) a non-commutative
composition rule that combines per-token updates, and (iii) a projection
operator that maps the continuous hidden state to a symbolic element of
the target finite group \texttt{G}. Formally, at sequence position
\texttt{t},

\[ s_t = F(s_{t-1}, x_t), \quad \hat{y}_L = \pi(s_L) \in G. \]

where \texttt{s\_t} is the continuous-valued hidden state, \texttt{F} is
a learned per-token update map, and $\pi : S \to G$ is the
projection operator. Because the per-token update is composed by an
associative composition operation, the full state sequence may
equivalently be computed in a scan-parallel form

\[ u_t = \phi(x_t), \quad s_L = u_1 \odot u_2 \odot \ldots \odot u_L, \quad \hat{y}_L = \pi(s_L). \]

where $\odot$ is an associative but \textbf{non-commutative}
composition operation. Order is preserved; sequential depth reduces to
$O(\log L)$.

\subsection{Hard projection and soft projection}\label{hard-projection-and-soft-projection}

The projection operator $\pi$ admits two regimes:

\begin{itemize}
\tightlist
\item
  \textbf{Hard projection}
  $\pi_{\text{hard}}(s) = \arg\min_{g \in G} d(s, \iota(g))$
  returns the unique nearest finite-group representative under a
  specified distance, where $\iota : G \to R$ embeds elements of
  \texttt{G} into a representation space \texttt{R}.
\item
  \textbf{Soft projection} with temperature
  $T > 0$ produces either a probability
  distribution
  $p_T(g | s) \propto \exp(-d(s, \iota(g)) / T)$
  over elements of \texttt{G}, or a convex combination of embedded
  representatives
  $\sum_g p_T(g|s) \cdot \iota(g)$.
\end{itemize}

As $T \to 0$, soft projection approaches hard projection. As
$T \to \infty$, it approaches a uniform mixture. The two regimes are
reported separately throughout this paper. Hard-projected outputs are
symbolic group elements; soft-projected outputs are distributions over
\texttt{G}.

\subsection{Public interface and implementation boundary}\label{public-interface-and-implementation-boundary}

This paper specifies the public interface needed to define the
benchmark, run the held-out transition-pair falsifier, interpret the
projected finite-group readout, and reproduce the diagnostic
calculations. The model interface consists of a continuous recurrent
state, an associative order-sensitive composition rule, and a projection
operator $\pi : S \to G$ that maps the final state to a symbolic element
of the target finite group.

The exact continuous carrier, the internal carrier embedding, and the
carrier-level constraint function used by the hard-projected model are
not part of this technical preprint. The empirical claims in this paper
should therefore be read as claims about the projected state-tracking
interface under the stated falsifier protocol, not as a full disclosure
of the carrier construction. The carrier-level construction is treated
separately from the protocol and readout contribution studied here.

\begin{center}\rule{0.5\linewidth}{0.5pt}\end{center}

\section{Experimental Protocol}\label{experimental-protocol}

\subsection{Gate A: baseline competence under matched protocol}\label{gate-a-baseline-competence-under-matched-protocol}

\textbf{Purpose.} To establish that the baseline models used elsewhere
in the paper are not generically incapable of solving easy
state-tracking tasks under the present protocol. Without this gate, any
subsequent baseline failure on the non-commutative task could be
attributed to under-trained or mis-configured baselines.

\textbf{Configuration.} Three baseline models are trained under matched
protocol: a bag-of-tokens model with continuous native readout, a Gated
Recurrent Unit (GRU) with continuous native readout, and a
single-configuration structured state-space model with continuous native
readout. The tasks are (i)
\texttt{easy commutative} and (ii)
\texttt{6-class noncommutative}. Training uses
$n_{\text{train}} = 1000$,
$n_{\text{val}} = 200$,
$n_{\text{test}} = 200$,
\texttt{epochs\ =\ 8}, \texttt{batch\_size\ =\ 64},
\texttt{lr\ =\ 0.003}, and five seeds
\texttt{\{20260525,\ 20260526,\ 20260527,\ 20260528,\ 20260529\}}. The
structured state-space model is run in a single configuration
($d_{\text{model}} = 64$,
$n_{\text{layers}} = 2$,
$d_{\text{state}} = 16$,
$d_{\text{conv}} = 4$,
$\text{expand} = 2$).

A disjoint-template overlap audit verifies zero verbatim-token,
template-ID, and template-family overlap across the easy-task and
held-out-task partitions used for each baseline.

\textbf{Gate A is a separate 6-class diagnostic control used to verify
baseline competence; it is not the $S_3 \times S_3$ held-out-pair
task reported in Gate B.} Its hard task is a 6-class non-commutative
final-state diagnostic chosen to be informative about ordinary
baselines' competence at order-sensitive prediction; its easy task is a
matched 6-class commutative control.

\subsection{Gate B: long-horizon held-out transition-pair performance}\label{gate-b-long-horizon-held-out-transition-pair-performance}

\textbf{Purpose.} The main empirical claim of this paper.

\textbf{Configuration.} Both the proposed projected-readout model and
the three baselines are trained on $S_3 \times S_3$ update
sequences under the held-out transition-pair split (Section 3.3), with
$n_{\text{train}} = 200$,
$n_{\text{val}} = 80$,
$n_{\text{test}} = 50$ per seed
(\texttt{250} total per horizon),
\texttt{train\_seq\_len\ =\ 8},
\texttt{epochs\ =\ 5}, \texttt{batch\_size\ =\ 40}, and
five seeds:
\texttt{\{20260525,\ 20260526,\ 20260527,\ 20260528,\ 20260529\}}.
Evaluation is conducted at horizons

\[ L_{\text{eval}} \in \{4096, 16384, 65536, 524288, 1048576\}. \]

With $L_{\text{train}} = 8$, the longest
evaluation horizon yields a length extrapolation ratio of
$L_{\text{eval}} / L_{\text{train}} = 131072 \geq 10^5$.

For the projected-readout model, training is conducted at a small but
nonzero projection temperature; long-horizon evaluation is reported
under hard projection. Local action supervision over $\Sigma$ and
presentation supervision enforcing
$a^{|a|} = e$ for each generator are
applied during training.

\subsubsection{Projection-matched baselines}\label{projection-matched-baselines}

To control for the possibility that baseline degradation under the
present protocol is a readout artifact rather than an architecture
difference, we add three projection-matched baselines: a GRU, a
single-configuration structured state-space model (same configuration as
in Section 5.1), and a bag-of-tokens encoder, each equipped with a learned
prototype-projection readout over the 36 elements of
$G = S_3 \times S_3$. The prototype readout is trained with
final group-state cross entropy only; no local-action or
presentation-supervision signals are added. The held-out transition-pair
split, seed list, training budget, evaluation horizons, and
$n_{\mathrm{test}}=50$ per seed match Section 5.2.

\subsection{Gate C: mechanism diagnostics under projection temperature}\label{gate-c-mechanism-diagnostics-under-projection-temperature}

\textbf{Purpose.} To probe whether the projected-readout model's
representation behaves approximately as a group homomorphism, and how
that behavior degrades as projection softens.

\textbf{Configuration.} Four diagnostics are evaluated across a sweep of
projection temperature

\[ T \in \{0.25, 0.50, 0.75, 1.00, 1.50, 2.00, 3.00\}. \]

at evaluation lengths
$\{2048, 8192\}$:

\begin{itemize}
\item
  \textbf{Final-token accuracy.} The probability that the model's
  prediction of \texttt{H\_L} equals the ground-truth group element.
\item
  \textbf{Exact homomorphism error.}
  $E_{(u, v)}\bigl{[} d\bigl( \pi(s(uv)), \pi(s(u)) \cdot_G \pi(s(v)) \bigr) \bigr{]}$
  over selected word pairs.
\item
  \textbf{State-consistency drift.} For a carrier admitting a constraint
  function \texttt{C(s)\ =\ 0} that an exact per-token update would
  preserve, we define state-consistency drift as

  \[ D = L^{-1} \sum_{t=1}^{L} \|C(s_t)\|. \]

  where \texttt{L} is the evaluation horizon, \texttt{s\_t} is the
  model's continuous-valued hidden state at position \texttt{t}, and
  $\|\cdot\|$
  is a specified matrix or vector norm. The functional form of
  \texttt{C} is a property of the carrier. The specific \texttt{C} used
  to compute the drift values reported in Section 6.3 is not disclosed
  in the present paper.
\item
  \textbf{Commutator gap.} For selected pairs
  $(x, y) \in \Sigma \times \Sigma$ whose group commutator
  \texttt{{[}x,\ y{]}} is non-identity, the distance between the model's
  representation of \texttt{{[}x,\ y{]}} and its representation of the
  group identity.
\end{itemize}

\subsection{Gate E: leakage and triviality firewall}\label{gate-e-leakage-and-triviality-firewall}

\textbf{Purpose.} To certify that the held-out evaluation partition does
not contain trivially leaked content from the training partition, and
that the perturbations the diagnostics evaluate are genuinely
non-commutative-specific.

\textbf{Configuration.} Five group-product specificity rates are
reported over the clean-split data structure:

\begin{itemize}
\tightlist
\item
  \texttt{contextual\_commutator}: rate at which deterministic
  commutator construction yields distinct group products in context;
\item
  \texttt{contextual\_inverse\_shuffle}: rate at which
  inverse-cancellation order perturbation changes the group product;
\item
  \texttt{held\_out\_generator\_pair}: rate at which paired held-out
  generator orders (\texttt{ab} vs \texttt{ba}) yield distinct group
  products;
\item
  \texttt{reversed\_word\_difference}: rate at which reversed-order
  perturbation changes the group product;
\item
  \texttt{same\_multiset\_different\_product}: rate at which two sequences
  with identical generator multisets yield distinct group products.
\end{itemize}

Each rate is reported with a bootstrap 95\% interval over five seeds. An
overlap audit additionally verifies zero verbatim reduced-word overlap
and zero structural-template overlap between training and evaluation
partitions.

\textbf{These rates are properties of the data generation process, not
direct neural-model accuracies.} Gate E is a leakage and triviality
firewall, not a model performance result.

\begin{center}\rule{0.5\linewidth}{0.5pt}\end{center}

\section{Results}\label{results}

\subsection{Gate A: baseline competence}\label{gate-a-baseline-competence}

{\def\LTcaptype{none} 
\begin{longtable}[]{@{}
  >{\raggedright\arraybackslash}p{(\linewidth - 8\tabcolsep) * \real{0.2600}}
  >{\raggedright\arraybackslash}p{(\linewidth - 8\tabcolsep) * \real{0.3000}}
  >{\raggedleft\arraybackslash}p{(\linewidth - 8\tabcolsep) * \real{0.0700}}
  >{\raggedleft\arraybackslash}p{(\linewidth - 8\tabcolsep) * \real{0.1400}}
  >{\raggedright\arraybackslash}p{(\linewidth - 8\tabcolsep) * \real{0.2300}}@{}}
\toprule\noalign{}
\begin{minipage}[b]{\linewidth}\raggedright
Model
\end{minipage} & \begin{minipage}[b]{\linewidth}\raggedright
Task
\end{minipage} & \begin{minipage}[b]{\linewidth}\raggedleft
Seeds
\end{minipage} & \begin{minipage}[b]{\linewidth}\raggedleft
Mean acc
\end{minipage} & \begin{minipage}[b]{\linewidth}\raggedright
95\% interval
\end{minipage} \\
\midrule\noalign{}
\endhead
\bottomrule\noalign{}
\endlastfoot
bag & easy commutative & 5 & 1.0000 &
{[}1.0000, 1.0000{]} \\
GRU & easy commutative & 5 & 1.0000 &
{[}1.0000, 1.0000{]} \\
structured SSM (single config) &
easy commutative & 5 & 0.8095 &
{[}0.7635, 0.8555{]} \\
bag & 6-class noncommutative & 5 & 0.1600 &
{[}0.1395, 0.1835{]} \\
GRU & 6-class noncommutative & 5 & 0.1690 &
{[}0.1355, 0.2015{]} \\
structured SSM (single config) &
6-class noncommutative & 5 & 0.1900 & {[}0.1380,
0.2650{]} \\
\end{longtable}
}

Chance accuracy on this 6-class diagnostic control is $1/6 \approx 0.1667$.

The bag and GRU baselines solve the easy commutative control at the
largest reported length with mean accuracy \texttt{1.0000}; the
single-configuration structured SSM solves it at \texttt{0.8095} $\pm$ seed
variation. On the 6-class non-commutative final-state diagnostic, all
three baselines remain at or near chance (\texttt{0.16}--\texttt{0.19}).
The baselines used in the present paper are therefore not generically
incompetent at state-tracking-style sequence tasks; they fail the
6-class non-commutative diagnostic under the same training protocol.

\begin{figure}
\centering
\includegraphics[width=0.90\linewidth,height=0.82\textheight,keepaspectratio]{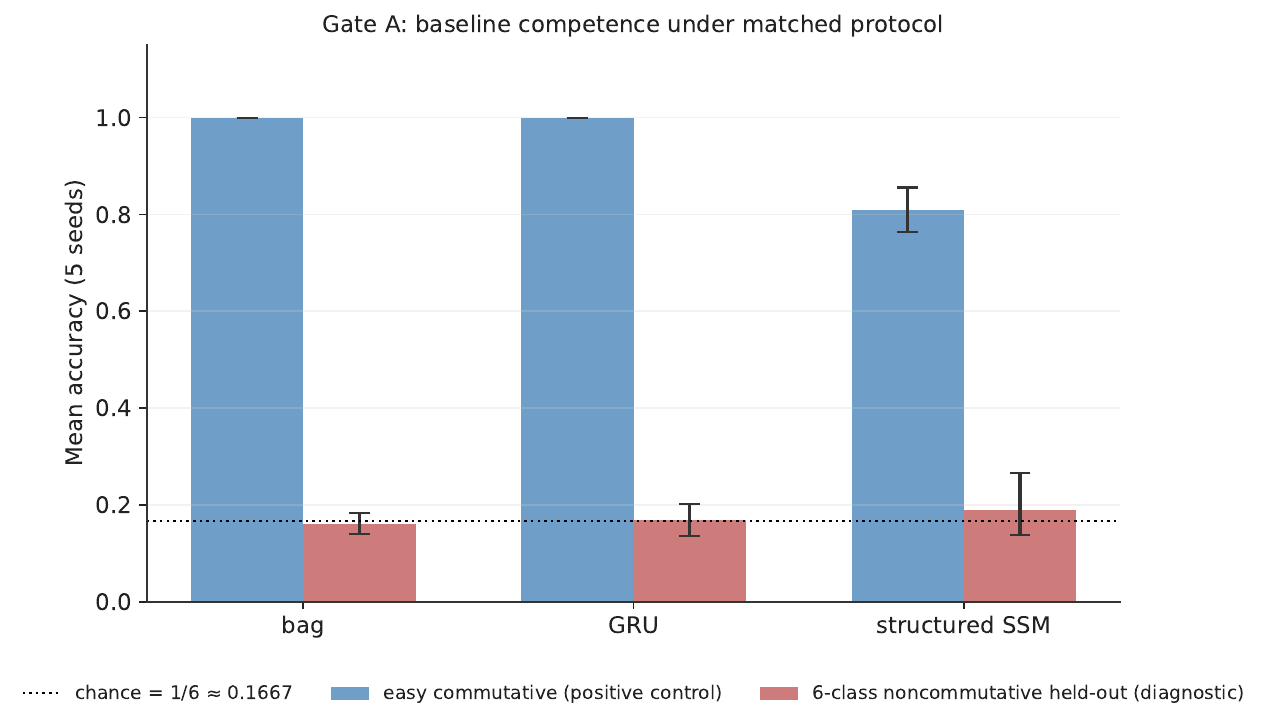}
\caption{Gate A baseline competence under matched protocol. The dotted line marks the $1/6$ chance level for the separate 6-class diagnostic control. Error bars indicate 95\% bootstrap intervals over five seeds.}
\label{fig:gate-a}
\end{figure}

\FloatBarrier
\subsection{Gate B: long-horizon held-out transition-pair performance}\label{gate-b-long-horizon-held-out-transition-pair-performance-1}

\textbf{Main result (expanded $n_{\mathrm{test}}=50$).}

{\def\LTcaptype{none} 
\begin{longtable}[]{@{}
  >{\raggedright\arraybackslash}p{(\linewidth - 12\tabcolsep) * \real{0.2050}}
  >{\raggedleft\arraybackslash}p{(\linewidth - 12\tabcolsep) * \real{0.1300}}
  >{\raggedleft\arraybackslash}p{(\linewidth - 12\tabcolsep) * \real{0.0750}}
  >{\raggedleft\arraybackslash}p{(\linewidth - 12\tabcolsep) * \real{0.0800}}
  >{\raggedleft\arraybackslash}p{(\linewidth - 12\tabcolsep) * \real{0.1400}}
  >{\raggedleft\arraybackslash}p{(\linewidth - 12\tabcolsep) * \real{0.1600}}
  >{\raggedleft\arraybackslash}p{(\linewidth - 12\tabcolsep) * \real{0.2100}}@{}}
\toprule\noalign{}
\begin{minipage}[b]{\linewidth}\raggedright
Model
\end{minipage} & \begin{minipage}[b]{\linewidth}\raggedleft
Eval length
\end{minipage} & \begin{minipage}[b]{\linewidth}\raggedleft
Seeds
\end{minipage} & \begin{minipage}[b]{\linewidth}\raggedleft
n per seed
\end{minipage} & \begin{minipage}[b]{\linewidth}\raggedleft
Exact / total
\end{minipage} & \begin{minipage}[b]{\linewidth}\raggedleft
Mean final acc
\end{minipage} & \begin{minipage}[b]{\linewidth}\raggedleft
95\% lower bound
\end{minipage} \\
\midrule\noalign{}
\endhead
\bottomrule\noalign{}
\endlastfoot
Hard-projected (ours) & 524288 & 5 & 50 & 250 / 250 & 1.0000 & 0.9854 \\
Hard-projected (ours) & 1048576 & 5 & 50 & 250 / 250 & 1.0000 &
0.9854 \\
\end{longtable}
}

The 95\% lower bound is a two-sided Clopper-Pearson interval lower
endpoint with \texttt{alpha\ =\ 0.05}.

\textbf{Short-horizon supplement (same $n_{\mathrm{test}}=50$
protocol).}

{\def\LTcaptype{none} 
\begin{longtable}[]{@{}
  >{\raggedright\arraybackslash}p{(\linewidth - 12\tabcolsep) * \real{0.2050}}
  >{\raggedleft\arraybackslash}p{(\linewidth - 12\tabcolsep) * \real{0.1300}}
  >{\raggedleft\arraybackslash}p{(\linewidth - 12\tabcolsep) * \real{0.0750}}
  >{\raggedleft\arraybackslash}p{(\linewidth - 12\tabcolsep) * \real{0.0800}}
  >{\raggedleft\arraybackslash}p{(\linewidth - 12\tabcolsep) * \real{0.1400}}
  >{\raggedleft\arraybackslash}p{(\linewidth - 12\tabcolsep) * \real{0.1600}}
  >{\raggedleft\arraybackslash}p{(\linewidth - 12\tabcolsep) * \real{0.2100}}@{}}
\toprule\noalign{}
\begin{minipage}[b]{\linewidth}\raggedright
Model
\end{minipage} & \begin{minipage}[b]{\linewidth}\raggedleft
Eval length
\end{minipage} & \begin{minipage}[b]{\linewidth}\raggedleft
Seeds
\end{minipage} & \begin{minipage}[b]{\linewidth}\raggedleft
n per seed
\end{minipage} & \begin{minipage}[b]{\linewidth}\raggedleft
Exact / total
\end{minipage} & \begin{minipage}[b]{\linewidth}\raggedleft
Mean final acc
\end{minipage} & \begin{minipage}[b]{\linewidth}\raggedleft
95\% lower bound
\end{minipage} \\
\midrule\noalign{}
\endhead
\bottomrule\noalign{}
\endlastfoot
Hard-projected (ours) & 4096 & 5 & 50 & 250 / 250 & 1.0000 & 0.9854 \\
Hard-projected (ours) & 16384 & 5 & 50 & 250 / 250 & 1.0000 & 0.9854 \\
Hard-projected (ours) & 65536 & 5 & 50 & 250 / 250 & 1.0000 & 0.9854 \\
\end{longtable}
}

\textbf{Statistical reporting note.} The expanded Gate B run uses 50
evaluations per seed across five fixed seeds (\texttt{250} samples per
horizon). All \texttt{250} evaluations at both \texttt{524288} and
\texttt{1048576} are error-free (perfect 250/250 at each horizon).
The two-sided Clopper-Pearson 95\% lower bound is \texttt{0.9854} at
both horizons. The earlier
\texttt{n\_test\ =\ 8} pilot evidence is superseded by this expanded
sampling. Broader sampling at the longest horizons remains future work.

\textbf{Soft / unprojected regime.} The error-free behavior under hard
projection must not be conflated with soft or unprojected behavior of
the same model. Under
soft evaluation at \texttt{T\ =\ 1.0} on the same evaluation set, the
model produces all-token accuracy $\approx 0.23$
/ $\approx 0.13$ at \texttt{524288} /
\texttt{1048576} and final-token accuracy \texttt{0.0000}. The two
regimes correspond to the two projection forms in Section 4.2 and are
reported separately throughout.

The hard-projected model produces error-free final-state predictions
(250/250 at each horizon) across two evaluation horizons separated
from the training horizon by approximately four to five orders of
magnitude. The expanded sampling makes the main
long-horizon result substantially less sensitive to small-\texttt{n}
pilot variance than the earlier \texttt{n\_test\ =\ 8} pilot.

\begin{figure}
\centering
\includegraphics[width=0.90\linewidth,height=0.82\textheight,keepaspectratio]{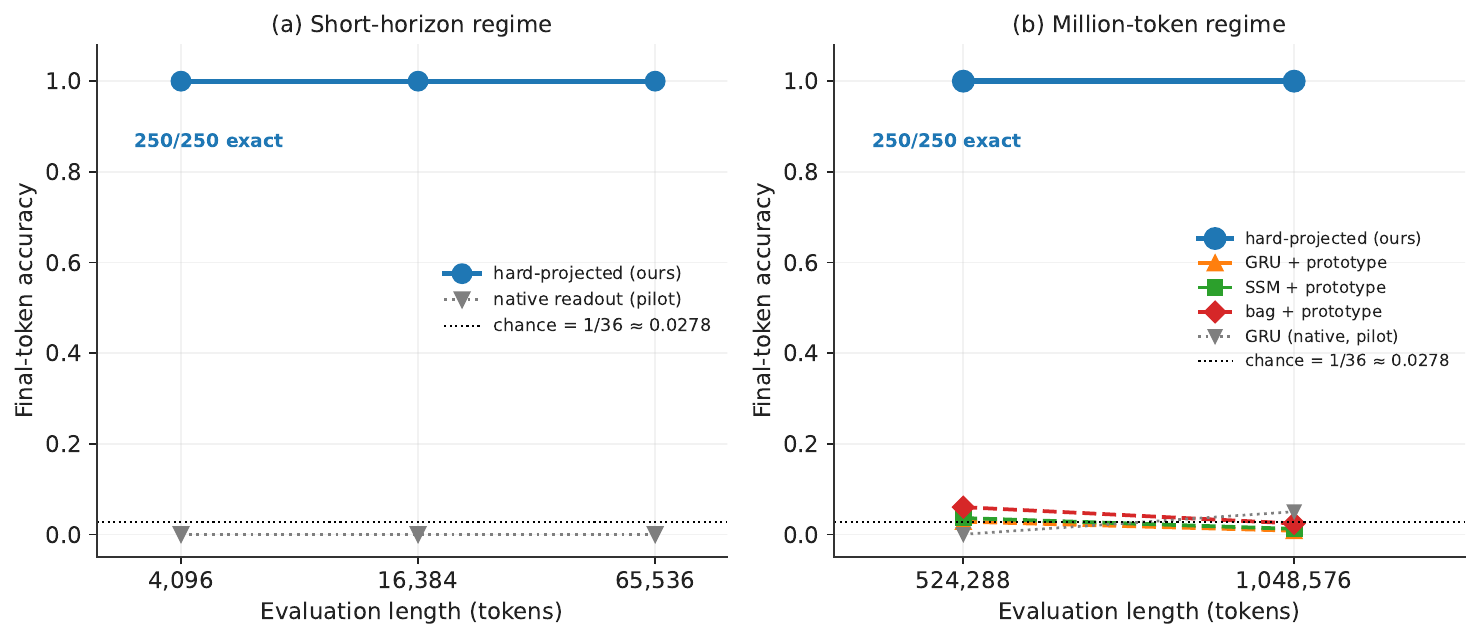}
\caption{Gate B held-out-pair falsifier. Panel (a) shows the short-horizon supplement. Panel (b) shows the million-token expanded test with projection-matched baselines (GRU, bag, structured SSM with prototype-projection readout) and the retained native-readout pilot reference. The dotted line marks chance accuracy for $S_3 \times S_3$, $1/36 \approx 0.0278$. Curves are shown only for model/horizon combinations implemented in the present harness.}
\label{fig:gate-b}
\end{figure}

\subsubsection{Native-readout baselines (retained from earlier pilot)}\label{native-readout-baselines-retained-from-earlier-pilot}

The native-readout baseline pilot from the earlier
\texttt{n\_test\ =\ 8} configuration is retained below as context. These
rows are not re-evaluated at $n_{\mathrm{test}}=50$ in this paper and
are reported only to preserve continuity with the original pilot.

{\def\LTcaptype{none} 
\begin{longtable}[]{@{}
  >{\raggedright\arraybackslash}p{(\linewidth - 10\tabcolsep) * \real{0.3450}}
  >{\raggedleft\arraybackslash}p{(\linewidth - 10\tabcolsep) * \real{0.1400}}
  >{\raggedleft\arraybackslash}p{(\linewidth - 10\tabcolsep) * \real{0.0750}}
  >{\raggedleft\arraybackslash}p{(\linewidth - 10\tabcolsep) * \real{0.0900}}
  >{\raggedleft\arraybackslash}p{(\linewidth - 10\tabcolsep) * \real{0.1700}}
  >{\raggedright\arraybackslash}p{(\linewidth - 10\tabcolsep) * \real{0.1800}}@{}}
\toprule\noalign{}
\begin{minipage}[b]{\linewidth}\raggedright
Model / regime
\end{minipage} & \begin{minipage}[b]{\linewidth}\raggedleft
Eval length
\end{minipage} & \begin{minipage}[b]{\linewidth}\raggedleft
Seeds
\end{minipage} & \begin{minipage}[b]{\linewidth}\raggedleft
n per seed
\end{minipage} & \begin{minipage}[b]{\linewidth}\raggedleft
Mean final acc
\end{minipage} & \begin{minipage}[b]{\linewidth}\raggedright
Read
\end{minipage} \\
\midrule\noalign{}
\endhead
\bottomrule\noalign{}
\endlastfoot
bag, native readout & 524288 & 5 & 8 & 0.0000 & pilot only \\
bag, native readout & 1048576 & 5 & 8 & 0.0000 & pilot only \\
GRU, native readout & 524288 & 5 & 8 & 0.0000 & pilot only \\
GRU, native readout & 1048576 & 5 & 8 & 0.0500 & pilot only \\
structured SSM, native readout & 4096 & 5 & 8 & 0.0000 & pilot only \\
structured SSM, native readout & 16384 & 5 & 8 & 0.0000 & pilot only \\
structured SSM, native readout & 65536 & 5 & 8 & 0.0000 & pilot only \\
\end{longtable}
}

\subsubsection{Projection-matched baselines}\label{projection-matched-baselines-1}

To test the readout-artifact hypothesis, three baselines with learned
prototype readout over the 36 elements of $G = S_3 \times S_3$
were trained and evaluated under the same protocol as Section 5.2 (held-out
transition-pair split, five fixed seeds, $n_{\mathrm{test}}=50$ per
seed, evaluation at \texttt{524288} and \texttt{1048576}). The chance
accuracy is $1/36 \approx 0.0278$.

{\def\LTcaptype{none} 
\begin{longtable}[]{@{}
  >{\raggedright\arraybackslash}p{(\linewidth - 10\tabcolsep) * \real{0.1800}}
  >{\raggedright\arraybackslash}p{(\linewidth - 10\tabcolsep) * \real{0.2400}}
  >{\raggedleft\arraybackslash}p{(\linewidth - 10\tabcolsep) * \real{0.1300}}
  >{\raggedleft\arraybackslash}p{(\linewidth - 10\tabcolsep) * \real{0.1300}}
  >{\raggedleft\arraybackslash}p{(\linewidth - 10\tabcolsep) * \real{0.1200}}
  >{\raggedleft\arraybackslash}p{(\linewidth - 10\tabcolsep) * \real{0.2000}}@{}}
\toprule\noalign{}
\begin{minipage}[b]{\linewidth}\raggedright
Model
\end{minipage} & \begin{minipage}[b]{\linewidth}\raggedright
Readout
\end{minipage} & \begin{minipage}[b]{\linewidth}\raggedleft
Eval length
\end{minipage} & \begin{minipage}[b]{\linewidth}\raggedleft
Exact / total
\end{minipage} & \begin{minipage}[b]{\linewidth}\raggedleft
Mean acc
\end{minipage} & \begin{minipage}[b]{\linewidth}\raggedleft
95\% lower bound
\end{minipage} \\
\midrule\noalign{}
\endhead
\bottomrule\noalign{}
\endlastfoot
GRU & prototype projection & 524288 & 7 / 250 & 0.0280 & 0.0113 \\
GRU & prototype projection & 1048576 & 2 / 250 & 0.0080 & 0.0010 \\
Structured SSM & prototype projection & 524288 & 9 / 250 & 0.0360 &
0.0166 \\
Structured SSM & prototype projection & 1048576 & 3 / 250 & 0.0120 &
0.0025 \\
Bag & prototype projection & 524288 & 15 / 250 & 0.0600 & 0.0340 \\
Bag & prototype projection & 1048576 & 6 / 250 & 0.0240 & 0.0089 \\
\end{longtable}
}

None of the projection-matched baselines approaches the hard-projected
result. Under the present tested configurations, the readout-artifact
hypothesis is not supported. We emphasise that this does not establish
exhaustive baseline impossibility; it tests fixed, pre-registered
baseline configurations under a matched held-out-pair protocol and
projection-readout control.

\begin{figure}
\centering
\includegraphics[width=0.90\linewidth,height=0.82\textheight,keepaspectratio]{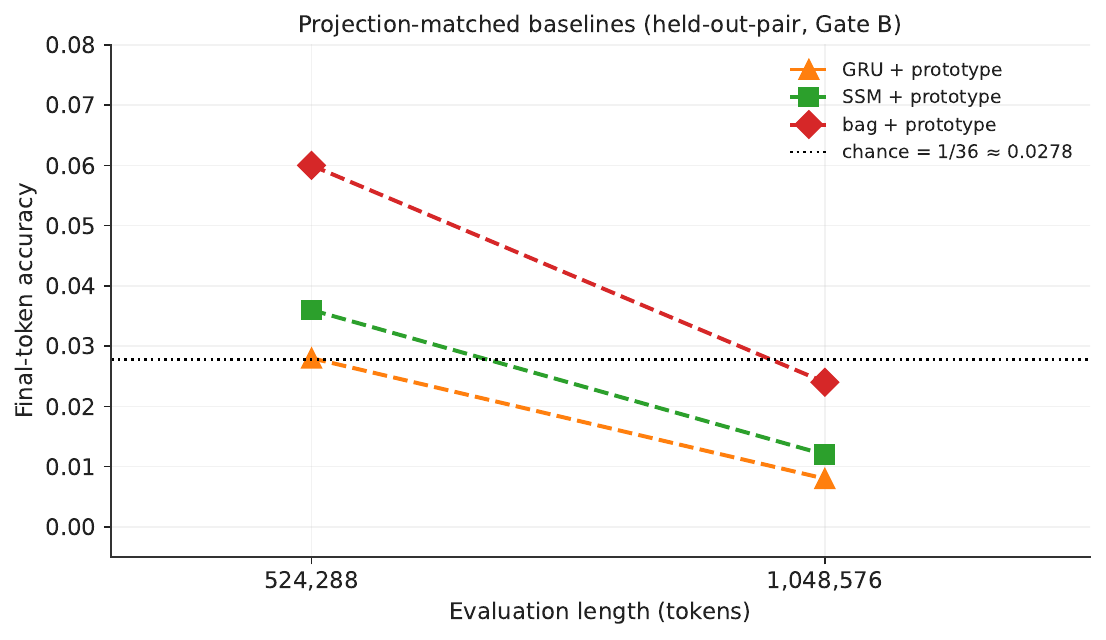}
\caption{Projection-matched baselines under the held-out-pair protocol at $L_{\mathrm{eval}} \in \{524{,}288, 1{,}048{,}576\}$. All baseline cells are far below the hard-projected $250/250$ result (all $\le 15/250$); the largest, bag at $524{,}288$ ($15/250$), is modestly above the $1/36$ chance line (dotted), while the rest sit at or near it.}
\label{fig:proj-matched}
\end{figure}

\subsection{Gate C: mechanism diagnostics across projection temperature}\label{gate-c-mechanism-diagnostics-across-projection-temperature}

{\def\LTcaptype{none} 
\begin{longtable}[]{@{}
  >{\raggedleft\arraybackslash}p{(\linewidth - 8\tabcolsep) * \real{0.0700}}
  >{\raggedleft\arraybackslash}p{(\linewidth - 8\tabcolsep) * \real{0.1400}}
  >{\raggedleft\arraybackslash}p{(\linewidth - 8\tabcolsep) * \real{0.2500}}
  >{\raggedleft\arraybackslash}p{(\linewidth - 8\tabcolsep) * \real{0.3100}}
  >{\raggedleft\arraybackslash}p{(\linewidth - 8\tabcolsep) * \real{0.2300}}@{}}
\toprule\noalign{}
\begin{minipage}[b]{\linewidth}\raggedleft
$T$
\end{minipage} & \begin{minipage}[b]{\linewidth}\raggedleft
Final acc
\end{minipage} & \begin{minipage}[b]{\linewidth}\raggedleft
Homomorphism error
\end{minipage} & \begin{minipage}[b]{\linewidth}\raggedleft
State-consistency drift
\end{minipage} & \begin{minipage}[b]{\linewidth}\raggedleft
Commutator gap
\end{minipage} \\
\midrule\noalign{}
\endhead
\bottomrule\noalign{}
\endlastfoot
0.25 & 1.0000 & 0.000583 & 0.032646 & 8.4844 \\
0.50 & 0.0400 & 0.192048 & 0.830739 & 8.1760 \\
0.75 & 0.0200 & 1.225186 & 0.825461 & 6.6129 \\
1.00 & 0.0600 & 2.960337 & 0.824627 & 4.2619 \\
1.50 & 0.0600 & 5.006641 & 0.828726 & 1.1692 \\
2.00 & 0.0400 & 5.478391 & 0.831072 & 0.3160 \\
3.00 & 0.0200 & 5.661507 & 0.832476 & 0.0448 \\
\end{longtable}
}

We observe a coherent boundary behavior beginning at $T = 0.50$: final-token accuracy collapses from $1.0000$ at $T=0.25$ to $0.0400$; homomorphism error rises by more than two orders of magnitude over the same step; state-consistency drift rises sharply and then plateaus near $0.83$ across the softened regime; and the commutator gap decays progressively from $8.48$ to $0.045$ as temperature increases.

In the lowest-temperature regime reported in the Gate C sweep
(\texttt{T\ =\ 0.25}), which approximates the hard-projection regime
used in Gate B, the model's projected representation exhibits very low
homomorphism error and very low drift, while preserving a large
commutator gap. This is the joint behavior expected of an approximately
group-homomorphic representation: the projection of a product equals the
product of projections (low
$E_{\text{homo}}$); the continuous state
remains close to the carrier's constraint manifold under $C$ (low
drift); and non-commuting pairs are separated from the identity (large
commutator gap). All three diagnostics co-degrade as projection softens.
The coincidence of these signals, rather than any single one alone,
constitutes the mechanism-level evidence the diagnostic family is
designed to deliver.

\begin{figure}
\centering
\includegraphics[width=0.90\linewidth,height=0.82\textheight,keepaspectratio]{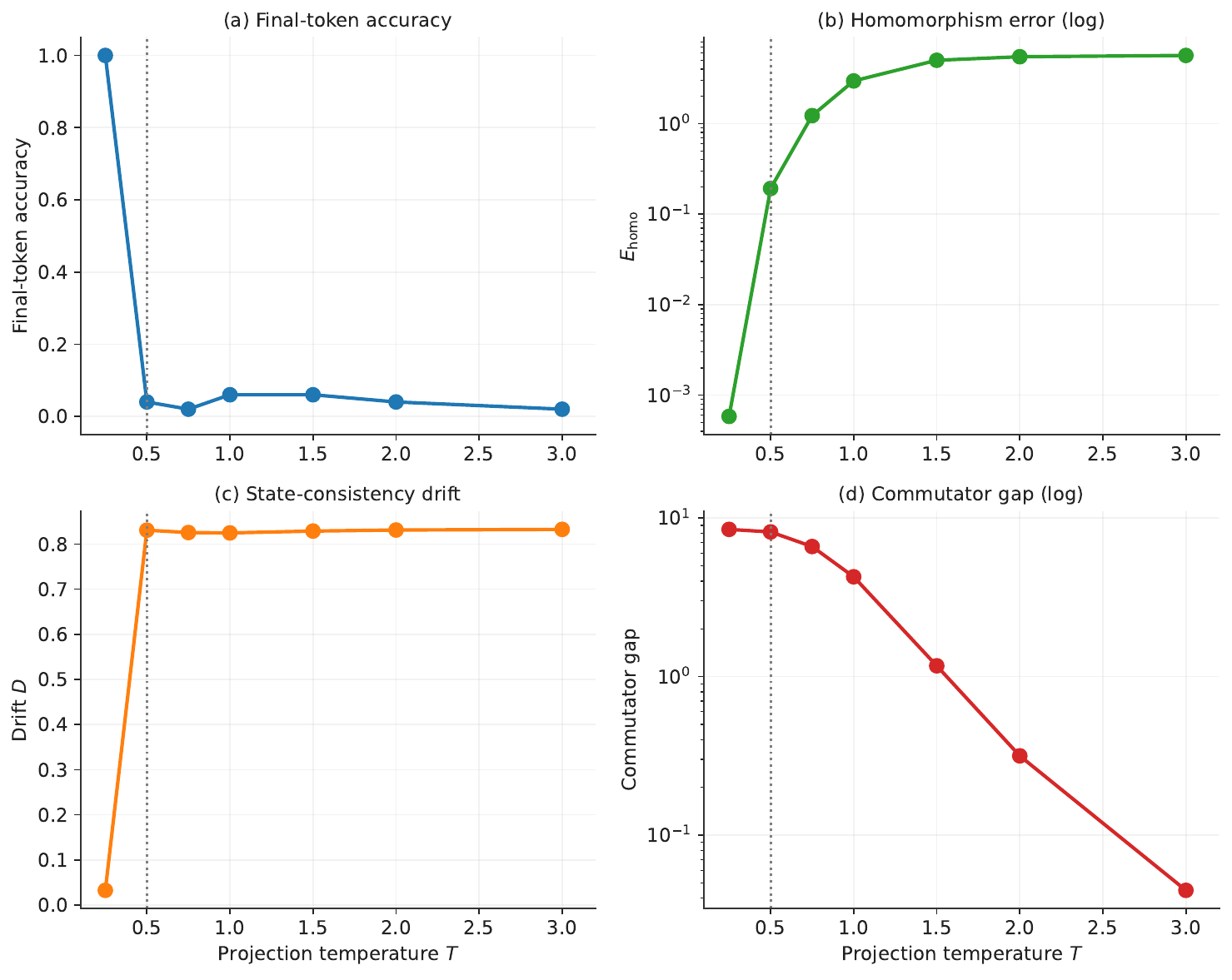}
\caption{Gate C mechanism diagnostics across projection temperature. Final-token accuracy collapses beginning at $T^\star=0.5$; homomorphism error and state-consistency drift rise; and commutator gap decays across the softened regime. Homomorphism error and commutator gap are shown on log scale. A dotted vertical line marks the boundary at $T^\star=0.5$ on all four panels.}
\label{fig:gate-c}
\end{figure}

\subsection{Gate E: leakage and triviality firewall}\label{gate-e-leakage-and-triviality-firewall-1}

{\def\LTcaptype{none} 
\begin{longtable}[]{@{}
  >{\raggedright\arraybackslash}p{(\linewidth - 4\tabcolsep) * \real{0.4800}}
  >{\raggedleft\arraybackslash}p{(\linewidth - 4\tabcolsep) * \real{0.2000}}
  >{\raggedleft\arraybackslash}p{(\linewidth - 4\tabcolsep) * \real{0.3200}}@{}}
\toprule\noalign{}
Specificity check & Mean (5 seeds) & Bootstrap 95\% interval \\
\midrule\noalign{}
\endhead
\bottomrule\noalign{}
\endlastfoot
contextual\_commutator & 1.0000 & {[}1.0000, 1.0000{]} \\
contextual\_inverse\_shuffle & 0.7936 & {[}0.7826, 0.8046{]} \\
held\_out\_generator\_pair & 0.3402 & {[}0.3248, 0.3556{]} \\
reversed\_word\_difference & 0.5554 & {[}0.5454, 0.5656{]} \\
same\_multiset\_different\_product & 1.0000 & {[}1.0000, 1.0000{]} \\
\end{longtable}
}

The clean-split overlap audit additionally verifies zero verbatim
reduced-word overlap and zero structural-template overlap between the
training and evaluation partitions of the held-out transition-pair
split.

The contextual commutator and same-multiset different-product checks
return $1.0000$: the data-generation process reliably produces inputs
in which commutator construction and same-multiset-different-product
reorderings yield distinct group products. Contextual inverse-shuffle
and reversed-word-difference return mid-range rates ($\approx 0.79$,
$\approx 0.56$), consistent with non-trivial but not maximal
sensitivity of these perturbations under the present construction. The
held-out generator-pair check returns $\approx 0.34$, reflecting the
proportion of contextualized constructions in which perturbing the
held-out generator order changes the full sequence product. This is a
contextual full-product check, not a claim that the adjacent two-token
products $(a_0, a_2)$ and $(a_2, a_0)$ are themselves distinct in the
direct-product convention. \textbf{None of these numbers is a
neural-model accuracy.} Gate E certifies that the held-out evaluation
partition contains genuinely non-commutative perturbations and is free
of trivial leakage.

\subsection{Robustness: same-factor held-out pair}\label{robustness-same-factor}

The held-out pair used for the main result, $\{(a_0,a_2),(a_2,a_0)\}$,
draws its two generators from different $S_3$ factors of the direct
product, which commute element-wise; the protocol therefore removes a
local transition \emph{template} rather than a directly non-commuting
adjacent pair (Section 3.3). To check that the result does not depend
on this cross-factor choice, we repeat the protocol with the held-out
pair drawn from within a single $S_3$ factor, where the two generators
genuinely do not commute. We run two such splits independently: a
first-factor split $\{(a_0,a_1),(a_1,a_0)\}$ and a second-factor split
$\{(a_2,a_3),(a_3,a_2)\}$, each with the same five seeds,
$n_{\mathrm{test}}=50$ per seed ($250$ total per horizon), training
length $L_{\mathrm{train}}=8$, and evaluation horizons
$L_{\mathrm{eval}} \in \{4096, 65536, 524288, 1048576\}$. For each split
a per-seed audit confirms that the held-out pair occurs zero times in
the training partition.

\begin{center}
\begin{tabular}{@{}llrrr@{}}
\toprule
Held-out split & Model & Exact / total & Mean acc & 95\% LB \\
\midrule
First factor & Hard-projected (ours) & 250 / 250 & 1.0000 & 0.9854 \\
First factor & GRU + projection & 2 / 250 & 0.0080 & 0.0010 \\
First factor & Structured SSM + projection & 3 / 250 & 0.0120 & 0.0025 \\
First factor & Bag + projection & 6 / 250 & 0.0240 & 0.0089 \\
\midrule
Second factor & Hard-projected (ours) & 250 / 250 & 1.0000 & 0.9854 \\
Second factor & GRU + projection & 3 / 250 & 0.0120 & 0.0025 \\
Second factor & Structured SSM + projection & 2 / 250 & 0.0080 & 0.0010 \\
Second factor & Bag + projection & 8 / 250 & 0.0320 & 0.0139 \\
\bottomrule
\end{tabular}
\end{center}

The hard-projected model is error-free ($250/250$) at every horizon for
both in-factor splits, identical to the cross-factor main result,
while the projection-matched GRU, structured SSM, and bag baselines all
remain near the $1/36$ chance line. The headline $1{,}048{,}576$-token
horizon is shown above; the $4096$, $65536$, and $524288$ horizons
exhibit the same pattern (hard-projected $250/250$ throughout; every
projection-matched baseline cell at or below $15/250$). The result is
therefore not an artifact of the cross-factor template choice: it holds
equally when the held-out adjacent pair is genuinely non-commuting, and
on either factor of the product group.

\begin{figure}
\centering
\includegraphics[width=0.90\linewidth,height=0.82\textheight,keepaspectratio]{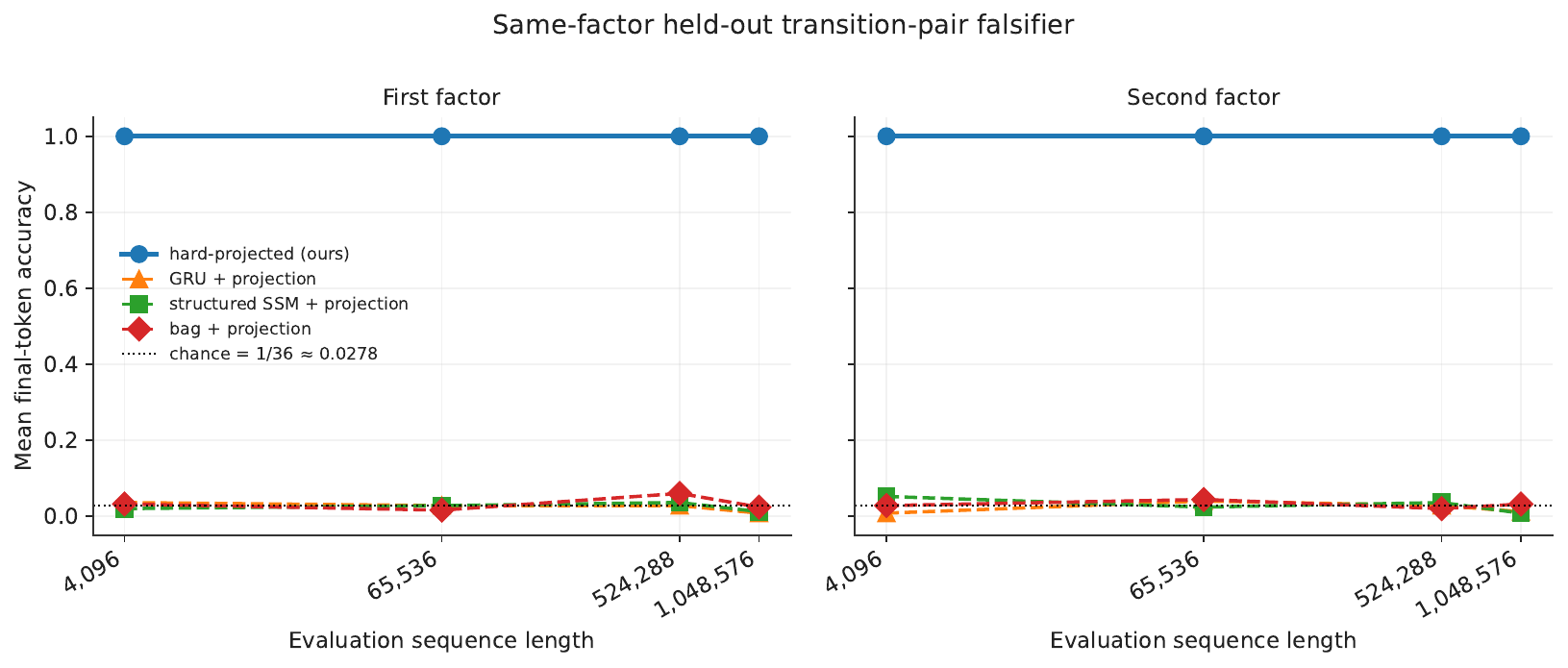}
\caption{Same-factor held-out robustness on $S_3 \times S_3$. Across both the first-factor and second-factor in-factor held-out splits, the hard-projected model is error-free at every evaluation horizon while the projection-matched GRU, structured SSM, and bag baselines remain near the $1/36$ chance line.}
\label{fig:same-factor}
\end{figure}

\begin{center}\rule{0.5\linewidth}{0.5pt}\end{center}

\section{Discussion}\label{discussion}

\subsection{What this demonstrates}\label{what-this-demonstrates}

Within a controlled state-tracking regime, an explicit projected
non-commutative state-composition interface provides a useful inductive
bias for preserving ordered hidden states across horizons orders of
magnitude longer than those seen in training. The held-out
transition-pair falsifier makes this claim precise by blocking one direct
local-transition memorization pathway. Under this protocol, the
projected-readout model maintains exact final-state accuracy at
million-token horizons, while matched baselines remain near chance. This
does not rule out every possible form of memorization or interpolation,
but it closes the most direct local-template pathway targeted by the
split.

\subsection{What this does not demonstrate}\label{what-this-does-not-demonstrate}

The result is deliberately narrow. It does not establish general
superiority of the proposed interface over Transformers, recurrent
networks, or state-space models on open-ended sequence tasks. It also
does not demonstrate that conventional sequence models are incapable of
non-commutative state composition in principle, only that the matched
baselines tested here fail under the specific constraints of this
falsifier. The present evidence is scoped to a controlled finite-group
benchmark with explicit projection; transfer to natural language, code,
or real-world workflows remains future work.

\subsection{Why this matters}\label{why-this-matters}

State-tracking evaluation in current practice does not always actively
block memorization pathways. As model families and benchmarks both grow,
the gap between ``appears to extrapolate'' and ``actually composes''
widens. We view the held-out transition-pair falsifier as a small but
reusable evaluation primitive that can be applied wherever a
sequence-modeling claim turns on order-sensitive hidden-state
composition. The projection-temperature diagnostic family is a
complementary primitive that distinguishes representations behaving
approximately as group homomorphisms from representations producing
correct outputs by other means.

\begin{center}\rule{0.5\linewidth}{0.5pt}\end{center}

\section{Limitations}\label{limitations}

A few limitations matter for how these results should be read.

\begin{enumerate}
\def\labelenumi{\arabic{enumi}.}
\item
  \textbf{Solvable target group.} The main result is on $S_3 \times S_3$.
  Appendix F reports a preliminary $S_5$ non-solvable stress test with
  both native-readout and projection-matched baselines, but it remains
  a single non-solvable group under a non-released carrier and carries its
  own carrier-embedding caveat; it is not a general non-solvable-group
  tracking result.
\item
  \textbf{Synthetic, controlled evaluation.} The benchmark is a
  finite-group state-tracking task. Real-world workflows, natural
  language, and code are out of scope, and no downstream application
  (e.g.\ workflow first-divergence localization or natural-language
  reasoning) is tested here.
\item
  \textbf{Hard projection is essential.} Exactness in Gate B holds under
  hard projection ($T \to 0$). Soft and unprojected variants of the
  same model collapse at the same horizons, and the two regimes are
  reported separately rather than being interchangeable.
\item
  \textbf{Baseline coverage is partial.} Projection-matched GRU,
  structured SSM, and bag baselines were run under the same held-out
  split, same seeds, and $n_{\mathrm{test}}=50$ on $S_3 \times S_3$
  (Section 6.2.2), on the two in-factor robustness splits
  (Section 6.5), and on $S_5$ (Appendix F). A broader sweep over
  state-space variants, attention-based architectures, and hybrids is
  left to future work. The projection-matched comparison does not by
  itself exhaust the readout-vs-architecture isolation question; it
  controls the readout artifact under fixed baseline configurations.
\item
  \textbf{No direct comparison to recent non-Abelian-state
  architectures under the same held-out split.} The Holonomic
  Network~{[}8{]}, PD-SSM~{[}6{]}, and M$^2$RNN~{[}7{]} are not run
  under the held-out transition-pair falsifier in this paper. A
  feasibility audit found that PD-SSM has public reference code but
  would require a custom data adapter and readout mapping for a fair
  matched comparison (deferred to a companion comparison), while no
  official reference implementation was located for M$^2$RNN or the
  Holonomic Network (literature-only positioning for now). Direct
  comparison therefore remains future work; the present evidence should
  be read as a protocol-and-readout result rather than as a model-level
  ordering over these architecture families.
\item
  \textbf{Carrier-level implementation boundary.} The continuous-carrier
  form of the projected recurrent state model is not described here. The
  benchmark, the falsifier, and the diagnostics can be reproduced without
  those details. Carrier construction is outside the scope of this
  protocol-and-readout preprint and is treated separately in an
  architecture report. The claim is not unsupervised discovery of the
  group law. The model uses local action supervision and
  presentation-level constraints; the evidence concerns the resulting
  projected state-composition interface under the held-out falsifier.
\end{enumerate}

Two further notes. The earlier pilot used $n_{\mathrm{test}}=8$ per
seed; the long-horizon evaluation reported here was expanded to
$n_{\mathrm{total}}=250$ per horizon on $S_3 \times S_3$, giving a
Clopper--Pearson $95\%$ lower bound of $0.9854$ under the tested
distribution. Broader sampling is still future work. This is a technical preprint: it prioritizes reproducibility of the
protocol, data generation, overlap audit, and diagnostics over end-to-end
model reproducibility.

\section{Conclusion}\label{conclusion}

We introduced a held-out transition-pair falsifier for non-Abelian state
tracking and showed that a projected recurrent state model preserves
exact final-state accuracy across million-token evaluation horizons
after short-horizon training, while matched native-readout baselines
remain near floor under the same protocol. Mechanism diagnostics across
a projection-temperature sweep exhibit a coherent boundary consistent
with approximately group-homomorphic behavior under hard projection.
The result is deliberately narrow. Its point is simple: when the state
is order, explicit projected non-commutative structure can be a useful
inductive bias for million-token horizons.

\appendix
\clearpage

\section{Full data generation protocol}\label{appendix-a-full-data-generation-protocol}

This appendix specifies the data generation procedure used in Gate B and
Gate E.

\begin{enumerate}
\def\labelenumi{\arabic{enumi}.}
\tightlist
\item
  \textbf{Group representation.} $S_3 \times S_3$ is represented
  as ordered pairs of \texttt{S\_3} elements under permutation-matrix
  arithmetic, with the group operation given componentwise.
\item
  \textbf{Generator definitions.}
  $\Sigma = \{a_0, a_1, a_2, a_3\}$,
  where \texttt{a\_0,\ a\_1} are two generators of the first
  \texttt{S\_3} factor and \texttt{a\_2,\ a\_3} are two generators of
  the second factor. Generators are selected such that each factor's
  generators together generate the full \texttt{S\_3}. Note that, by the
  direct-product structure, any generator from the first factor commutes
  with any generator from the second factor.
\item
  \textbf{Sequence generation.} Sequences are drawn by sampling
  generators independently from $\Sigma$ and then post-processing to
  (a) reject training sequences containing any forbidden ordered pair as
  consecutive generators, and (b) inject required ordered pairs into
  evaluation sequences at sampled positions.
\item
  \textbf{Rejection sampling for forbidden pairs.} Training sequences
  are sampled by independent generator draw, then any sequence
  containing a forbidden pair is rejected and resampled. Rejection rate
  at $L_{\text{train}} = 8$ is moderate and
  is recorded in the seed manifest.
\item
  \textbf{Required-pair insertion.} Evaluation sequences are sampled and
  then deterministically modified to include each required pair in at
  least one consecutive position; the modification is constructed to
  preserve the unconstrained nature of the remaining positions.
\end{enumerate}

A frozen seed manifest is emitted with each Gate B and Gate E rollup.

\section{Overlap audit details}\label{appendix-b-overlap-audit-details}

The clean-split overlap audit computes two overlap statistics between
the training and evaluation partitions:

\begin{itemize}
\tightlist
\item
  \textbf{Verbatim reduced-word overlap.} The fraction of evaluation
  reduced-word signatures that also appear verbatim in the set of
  training reduced-word signatures, where each sequence is reduced under
  the relations of the target group prior to comparison.
\item
  \textbf{Structural-template overlap.} The fraction of evaluation
  structural templates (length, generator-multiset, and ordered-pair
  multiset) that also appear in the training set under the same template
  criterion.
\end{itemize}

Both statistics are reported per check and per seed. The clean-split
data used in Gate E reports zero overlap under both criteria. A separate
audit run on the development data of an earlier prototype identified a
non-zero reduced-word overlap on three of five checks; that prototype is
not reported as the present paper's evidence, and was demoted to
structural-probe status prior to the clean-split rebuild.

\section{Baseline configurations}\label{appendix-c-baseline-configurations}

\begin{itemize}
\tightlist
\item
  \textbf{Bag-of-tokens baseline.} Per-token embeddings averaged across
  the sequence, followed by a linear continuous native readout over the
  class label set.
\item
  \textbf{Gated Recurrent Unit (GRU).} Standard single-layer or
  two-layer GRU, hidden size matched to baseline budget, with a
  continuous native readout.
\item
  \textbf{Structured state-space baseline.} Single configuration:
  $d_{\text{model}} = 64$,
  $n_{\text{layers}} = 2$,
  $d_{\text{state}} = 16$,
  $d_{\text{conv}} = 4$,
  $\text{expand} = 2$. Continuous native
  readout.
\end{itemize}

All baselines share training protocol,
$n_{\text{train}}$ /
$n_{\text{val}}$ /
$n_{\text{test}}$, optimizer, learning rate,
batch size, and seed list with the projected-readout model on each task.

\section{Prior empirical context (motivation only)}\label{appendix-d-prior-empirical-context-motivation-only}

Earlier internal experiments on related but distinct finite-group
tracking tasks (including permutation tasks on \texttt{S\_3},
\texttt{S\_5}, and several program-state benchmarks) motivated the
stricter held-out transition-pair falsifier introduced in this paper.
Those earlier results are not the main evidence reported here and are
not assumed by the present claims; they are referenced only as
motivation for the present protocol design.

\begin{center}\rule{0.5\linewidth}{0.5pt}\end{center}

\section{Public Interface of the Model}\label{appendix-e-minimal-interface-of-the-projected-recurrent-state-model}

This appendix specifies the public-facing interface of the projected
recurrent state model used in this paper, enough to follow Sections
4--6 without seeing the carrier-level implementation. The exact
carrier, embedding map, and constraint function are outside the scope
of this protocol-and-readout preprint.

\subsection*{E.1 Computation graph}

At sequence position $t$, the model maintains a continuous-valued
recurrent hidden state $s_t$ and computes:
\begin{itemize}
  \item input: a generator token $x_t \in \Sigma$;
  \item per-token update: $u_t = \varphi(x_t)$, where $\varphi$ is a
        learned map from $\Sigma$ into the continuous state space;
  \item associative non-commutative composition: $s_L = u_1 \odot u_2
        \odot \ldots \odot u_L$, equivalently realized by a recurrent
        accumulation $s_t = s_{t-1} \odot u_t$;
  \item projection: $\pi : S \to G$, mapping the continuous state to
        a symbolic element of the target finite group;
  \item output: $\hat{y}_L = \pi(s_L) \in G$, the predicted accumulated
        group element.
\end{itemize}

\subsection*{E.2 Training-signal composition}

The model is trained with the following loss components:
\begin{itemize}
  \item final-state cross entropy on the predicted symbolic group
        element $\hat{y}_L$ against the ground-truth $H_L$;
  \item local action supervision: for each generator $a \in \Sigma$, a
        target update on the continuous state consistent with the
        action of $a$ on the target group;
  \item presentation supervision: the defining relations of $G$ are
        imposed on the representation (e.g., $a^{|a|} = e$ for each
        generator $a$);
  \item projection/readout consistency, when used, regularises the
        agreement between hard and soft projection at low temperature.
\end{itemize}

\subsection*{E.3 Implementation boundary}

This appendix describes the public computation graph and training-signal
family used to interpret the reported experiments. It does not specify
the exact continuous carrier, internal embedding map, or carrier
constraint function. Those implementation details are outside the scope
of this protocol-and-readout preprint. The benchmark, the held-out
transition-pair falsifier, and the diagnostic family of Section 5.3 are
defined and reproducible independently of these carrier-level details.

\begin{center}\rule{0.5\linewidth}{0.5pt}\end{center}

\section{\texorpdfstring{Preliminary \texttt{S\_5} Non-Solvable Stress Evidence}{Preliminary S\_5 Non-Solvable Stress Evidence}}\label{appendix-f-preliminary-s_5-non-solvable-stress-evidence}

This appendix reports a preliminary stress test on the non-solvable group
\texttt{S\_5} ($|G| = 120$). It is included
as preliminary scope evidence and is not part of the main paper claim.

\subsection{Group setup and split}\label{f.1-group-setup-and-split}

The target group is the symmetric group \texttt{S\_5}. Generators
(Cayley): \texttt{r\ =\ (1\ 2\ 3\ 4\ 5)} of order 5 and
\texttt{s\ =\ (1\ 2)} of order 2. These generate \texttt{S\_5} and do
not commute. Generator set convention: positive only,
$\Sigma = \{r, s\}$. The held-out transition-pair split is
\texttt{P\_forbid\ =\ P\_require\ =\ \{(r,\ s),\ (s,\ r)\}}. Chance
accuracy for the full group-state classification is
$1/120 \approx 0.0083$.

\subsection{Configuration}\label{f.2-configuration}

Both a short-horizon configuration ($L_{\mathrm{train}}=8$,
$L_{\mathrm{eval}} \in \{512, 2048, 8192\}$) and an extended configuration
($L_{\mathrm{train}}=16$,
$L_{\mathrm{eval}} \in \{512, 2048, 8192, 65536\}$) were executed with
$n_{\mathrm{test}}=50$ per seed across the same five seeds. The
aggregate table below combines the two configurations for the three
shared shorter horizons.

\subsection{Results}\label{f.3-results}

{\def\LTcaptype{none} 
\begin{longtable}[]{@{}
  >{\raggedright\arraybackslash}p{(\linewidth - 10\tabcolsep) * \real{0.2900}}
  >{\raggedleft\arraybackslash}p{(\linewidth - 10\tabcolsep) * \real{0.1400}}
  >{\raggedleft\arraybackslash}p{(\linewidth - 10\tabcolsep) * \real{0.1600}}
  >{\raggedleft\arraybackslash}p{(\linewidth - 10\tabcolsep) * \real{0.1300}}
  >{\raggedleft\arraybackslash}p{(\linewidth - 10\tabcolsep) * \real{0.1900}}
  >{\raggedleft\arraybackslash}p{(\linewidth - 10\tabcolsep) * \real{0.0900}}@{}}
\toprule\noalign{}
\begin{minipage}[b]{\linewidth}\raggedright
Model
\end{minipage} & \begin{minipage}[b]{\linewidth}\raggedleft
Eval length
\end{minipage} & \begin{minipage}[b]{\linewidth}\raggedleft
Exact / total
\end{minipage} & \begin{minipage}[b]{\linewidth}\raggedleft
Mean acc
\end{minipage} & \begin{minipage}[b]{\linewidth}\raggedleft
95\% lower bound
\end{minipage} & \begin{minipage}[b]{\linewidth}\raggedleft
Chance
\end{minipage} \\
\midrule\noalign{}
\endhead
\bottomrule\noalign{}
\endlastfoot
Hard-projected (ours) & 512 & 500 / 500 & 1.0000 & 0.9926 & 0.0083 \\
Hard-projected (ours) & 2048 & 500 / 500 & 1.0000 & 0.9926 & 0.0083 \\
Hard-projected (ours) & 8192 & 500 / 500 & 1.0000 & 0.9926 & 0.0083 \\
Hard-projected (ours) & 65536 & 250 / 250 & 1.0000 & 0.9854 & 0.0083 \\
GRU native readout & 512 & 3 / 500 & 0.0060 & 0.0012 & 0.0083 \\
GRU native readout & 2048 & 4 / 500 & 0.0080 & 0.0022 & 0.0083 \\
GRU native readout & 8192 & 3 / 500 & 0.0060 & 0.0012 & 0.0083 \\
GRU native readout & 65536 & 0 / 250 & 0.0000 & 0.0000 & 0.0083 \\
\end{longtable}
}

\subsubsection*{Projection-matched baselines on \texorpdfstring{$S_5$}{S\_5}}

To remove the readout as a confound on this stress setting, we
additionally ran prototype-projection GRU, structured SSM, and bag
baselines over the $120$ elements of $S_5$ under the same held-out
split, the same five seeds, and $n_{\mathrm{test}}=50$ per seed
($250$ total per horizon). All three projection-matched baselines
remain near the $1/120 \approx 0.0083$ chance reference; the largest
single aggregate cell is bag $+$ projection at $3/250$.

\begin{center}
\begin{tabular}{@{}lrrr@{}}
\toprule
Model & Eval length & Exact / total & Mean acc \\
\midrule
GRU + projection & 512 & 0 / 250 & 0.0000 \\
GRU + projection & 2048 & 1 / 250 & 0.0040 \\
GRU + projection & 8192 & 1 / 250 & 0.0040 \\
GRU + projection & 65536 & 0 / 250 & 0.0000 \\
Structured SSM + projection & 512 & 0 / 250 & 0.0000 \\
Structured SSM + projection & 2048 & 1 / 250 & 0.0040 \\
Structured SSM + projection & 8192 & 1 / 250 & 0.0040 \\
Structured SSM + projection & 65536 & 1 / 250 & 0.0040 \\
Bag + projection & 512 & 2 / 250 & 0.0080 \\
Bag + projection & 2048 & 3 / 250 & 0.0120 \\
Bag + projection & 8192 & 1 / 250 & 0.0040 \\
Bag + projection & 65536 & 2 / 250 & 0.0080 \\
\bottomrule
\end{tabular}
\end{center}

\begin{figure}
\centering
\includegraphics[width=0.90\linewidth,height=0.82\textheight,keepaspectratio]{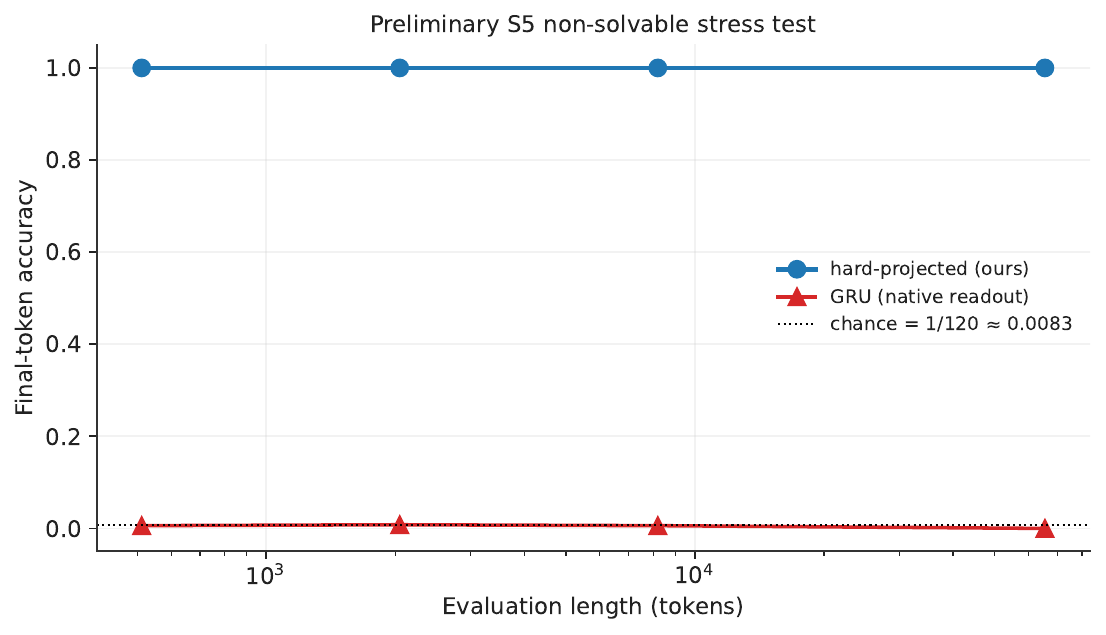}
\caption{Preliminary $S_5$ non-solvable stress test. The hard-projected model remains exact across the executed horizons, while native-readout GRU remains near chance. Chance line at $1/120 \approx 0.0083$.}
\label{fig:s5-stress}
\end{figure}

\begin{figure}
\centering
\includegraphics[width=0.90\linewidth,height=0.82\textheight,keepaspectratio]{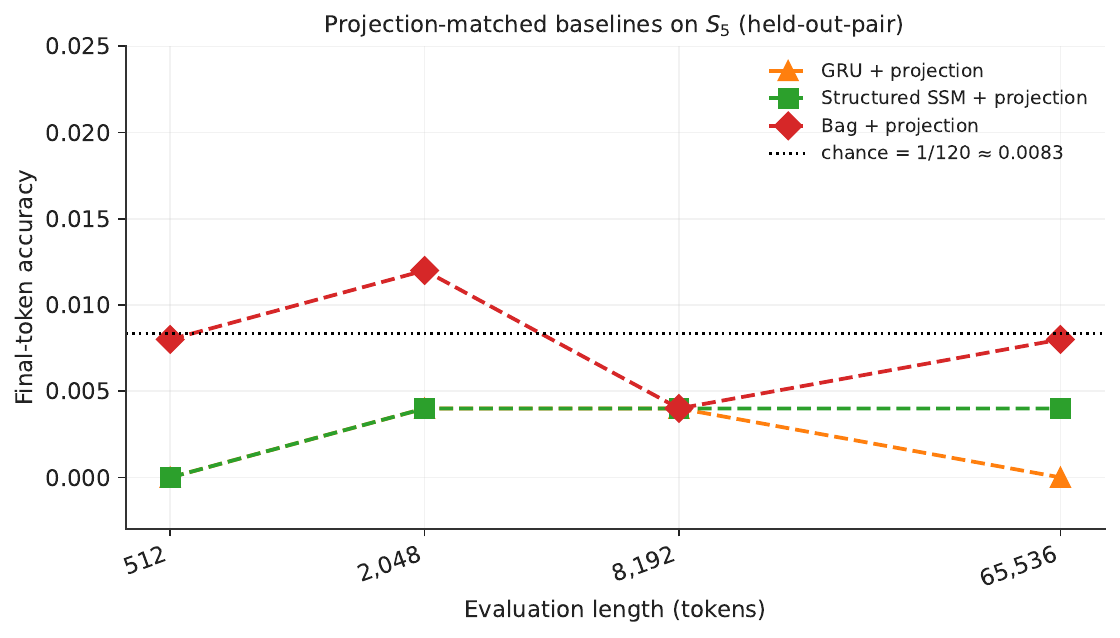}
\caption{Projection-matched baselines on $S_5$ under the held-out-pair protocol. Prototype-projection GRU, structured SSM, and bag baselines all remain near the $1/120 \approx 0.0083$ chance line, mirroring the native-readout GRU and supporting that the hard-projected $S_5$ result is not explained by the tested projection-readout artifact hypothesis.}
\label{fig:s5-proj-matched}
\end{figure}

\subsection{Honest caveat}\label{f.4-honest-caveat}

Perfect \texttt{S\_5} accuracy under the present projected-readout
interface raises a natural architectural question: does the non-released
carrier instantiate or approximate a representation of \texttt{S\_5} in
a way that makes this stress setting especially favourable? That mechanism analysis is outside the scope of the present preprint. The present empirical claim is bounded by the protocol: the
hard-projected model solves the reported held-out split on \texttt{S\_5}
under the stated configuration, while a GRU with native readout remains
near chance.

\textbf{This appendix should not be read as evidence that arbitrary
non-solvable groups are solved by the interface; it is a single stress
test under a non-released implementation.} The result should not be framed
as a broad complexity-class separation or a universal claim about all
sequence models. Both native-readout and projection-matched baselines
(GRU, structured SSM, bag) remain near chance on this split, so the gap
is not a readout artifact under the tested configurations; but this
remains a single non-solvable group under a non-released carrier, not a
general non-solvable-tracking result.

\subsection{Wall-clock summary}\label{f.5-wall-clock-summary}

Per-seed evaluation wall-clock times for the hard-projected row were
short and broadly comparable to the GRU baseline (illustrative median
values: approximately 0.03 s at \texttt{512}, 0.10 s at \texttt{2048},
0.41 s at \texttt{8192}, and 3.3 s at \texttt{65536} on the reported
CUDA device). A more thorough timing study is left to future work.

\section{Reproducibility notes}\label{appendix-g-reproducibility-notes}

\subsection{Eval-set provenance}\label{g.1-eval-set-provenance}

For the main $S_3 \times S_3$ result, the runner uses the same
deterministic generator call for every phase, and recomputed
token+label SHA-256 hashes are identical across phases for each
\texttt{(seed,\ eval\_length)} cell; the audit certifies identity by
deterministic regeneration. For the same-factor robustness splits
(Section 6.5) and the $S_5$ projection-matched runs (Appendix F), the
actual evaluation token and label arrays are additionally persisted as
files, each with a recorded SHA-256, a token-encoding specification,
and required-pair insertion positions, supporting independent
byte-for-byte replay without regeneration. A separate per-seed audit
of the length-8 training partitions confirms zero occurrences of the
held-out pair in training for the main split, the two in-factor splits,
and the $S_5$ split.

\subsection{Structured-SSM dependency note}\label{g.2-structured-ssm-dependency-note}

The structured SSM baseline used
\texttt{mamba-ssm\ 2.3.2.post1+cu11torch2.6} on an NVIDIA A100 device. A
small compatibility shim (\texttt{triton.set\_allocator} no-op) was
added at the import site to bypass an attribute the installed Triton
package did not expose. The shim affects only import/runtime
compatibility; it does not alter data, labels, split policy, or reported
outcomes. A version manifest is maintained alongside the artifacts.

\section*{References}\label{references}

{[}1{]} David A. Barrington. \emph{Bounded-Width Polynomial-Size
Branching Programs Recognize Exactly Those Languages in NC\^{}1}.
Journal of Computer and System Sciences, 38(1):150--164, 1989.
doi:10.1016/0022-0000(89)90037-8.

{[}2{]} Kenneth Krohn and John Rhodes. \emph{Algebraic Theory of
Machines. I. Prime Decomposition Theorem for Finite Semigroups and
Machines}. Transactions of the American Mathematical Society,
116:450--464, 1965. doi:10.2307/1994127.

{[}3{]} William Merrill, Jackson Petty, and Ashish Sabharwal. \emph{The
Illusion of State in State-Space Models}. ICML 2024; arXiv:2404.08819,
2024.

{[}4{]} Mehran Shakerinava, Behnoush Khavari, Siamak Ravanbakhsh, and
Sarath Chandar. \emph{The Expressive Limits of Diagonal SSMs for
State-Tracking}. arXiv:2603.01959, 2026.

{[}5{]} M. Reza Ebrahimi, Michaël Defferrard, Sunny Panchal, and Roland
Memisevic. \emph{On the ``Induction Bias'' in Sequence Models}.
arXiv:2602.18333, 2026.

{[}6{]} Aleksandar Terzić, Nicolas Menet, Michael Hersche, Thomas
Hofmann, and Abbas Rahimi. \emph{Structured Sparse Transition Matrices
to Enable State Tracking in State-Space Models}. NeurIPS 2025 Spotlight;
arXiv:2509.22284, 2025.

{[}7{]} Mayank Mishra, Shawn Tan, Ion Stoica, Joseph Gonzalez, and Tri
Dao. \emph{$M^2$RNN: Non-Linear RNNs with Matrix-Valued States for Scalable
Language Modeling}. arXiv:2603.14360, 2026.

{[}8{]} Ilmo Sung. \emph{Robust Reasoning as a Symmetry-Protected
Topological Phase}. arXiv:2601.05240, 2026.

\begin{center}\rule{0.5\linewidth}{0.5pt}\end{center}

\section*{Code and Data Availability}\label{code-and-data-availability}

The public release package includes benchmark generation code, held-out
split construction, overlap-audit scripts, Gate~E specificity checks,
result CSVs, figure scripts, evaluation-set hashes, and
projection-matched baseline configurations. These artifacts are intended
to make the falsifier protocol, clean-split audit, baseline controls,
and reported figures independently inspectable.

The code, configurations, figures, result tables, and evaluation-set
hashes are available at
\url{https://github.com/jeonghoon-ad/heldout-transition-pair-falsifier}.
The long-horizon evaluation token/label arrays (524{,}288 and
1{,}048{,}576 tokens) are archived at
\url{https://doi.org/10.5281/zenodo.20506128}.

\begin{center}
\small
\begin{tabular}{lll}
\hline
Component & Released? & Purpose \\
\hline
Benchmark generator & Yes & Reproduce held-out splits \\
Held-out split manifests & Yes & Verify train/eval construction \\
Overlap-audit scripts & Yes & Check leakage / structural overlap \\
Gate E specificity audit & Yes & Reproduce data-level firewall checks \\
Result CSVs & Yes & Reproduce reported tables \\
Figure scripts & Yes & Reproduce reported plots \\
Proj.-matched baseline configs & Yes & Reproduce baseline controls \\
Evaluation-set hashes & Yes & Verify provenance of evaluations \\
Hard-projected model carrier & No & Outside this technical preprint \\
Carrier constraint function & No & Outside this technical preprint \\
Core model training code & No & Not in the public release \\
\hline
\end{tabular}
\end{center}

The release does not include the non-released carrier implementation
used by the hard-projected model. Full end-to-end reproduction of that
implementation is outside the scope of this technical preprint (see
Section~8). Reproducibility notes on the structured-SSM dependency are
listed in Appendix G.

Some methods described in this paper are the subject of a pending patent
application filed by the author.

\begin{center}\rule{0.5\linewidth}{0.5pt}\end{center}

\section*{Acknowledgements}\label{acknowledgements}

The author acknowledges the broader theoretical work on state-tracking
expressivity by Merrill, Petty, Sabharwal, Shakerinava, Ebrahimi, and
collaborators referenced above, whose work framed the question that this
paper attempts a narrow constructive empirical response to.

AI tools were used for technical exploration, code development, data
analysis, figure preparation, and manuscript writing and editing.

\end{document}